\documentclass{llncs}
\usepackage[utf8]{inputenc}
\usepackage{makeidx}  
\usepackage{listings}

\usepackage[T1]{fontenc}
\usepackage{lmodern}
\usepackage[scaled=0.7]{beramono}
\usepackage{amsfonts}
\usepackage{amsmath}
\usepackage{url}
\usepackage{graphicx}
\usepackage{soul}
\usepackage{color}
\usepackage{multirow}
\usepackage{xspace}
\usepackage{stmaryrd}
\usepackage{array}
\usepackage[draft]{fixme}
\usepackage{wrapfig}
\usepackage{subfig}
\usepackage[font=scriptsize,labelfont=bf]{caption}
\lstdefinelanguage{SPARQLStream}[]{SQL}{
                        morekeywords={RANGE,FILTER,RSTREAM,ISTREAM,DSTREAM,TO,SLIDE,NAMED,
                        STREAM,NOW,PREFIX,STEP,REGISTER, QUERY}}
\lstdefinestyle{SPARQLStreamStyle}{basicstyle=\ttfamily\small,keywordstyle=\bfseries,
                        emphstyle=\itshape,showstringspaces=false,
                        frame=single,captionpos=b}

\lstdefinestyle{XMLStyle}{basicstyle=\ttfamily\scriptsize,keywordstyle=\bfseries,
                        emphstyle=\itshape,showstringspaces=false,
                        frame=single,captionpos=b}

\begin{document}
\captionsetup{format=plain}
\frontmatter          

\mainmatter              
\title{The Schema Editor of OpenIoT for Semantic Sensor Networks }
\titlerunning{}  
%
\vspace{-10pt}

\author{Prem Prakash Jayaraman\inst{1} \and Jean-Paul Calbimonte\inst{2}
\and Hoan Nguyen Mau Quoc\inst{3}}
\authorrunning{} 
%
%
\vspace{-5pt}

\institute{RMIT University, Melbourne, Australia.\\
\email{prem.jayaraman@rmit.edu.au}
\and
EPFL, Switzerland\\
\email{\{name.surname\}@epfl.ch}
\and
Insight Centre for Data Analytics, National University of Ireland, Galway, Ireland\\
\email{hoan.quoc@insight-centre.org}
}


\maketitle              

\vspace{-15pt}

\begin{abstract}
Ontologies provide conceptual abstractions over data, in domains such as the Internet of Things, in a way that sensor data can be harvested and interpreted by people and applications. The Semantic Sensor Network (SSN) ontology is the de-facto standard for semantic representation of sensor observations and metadata, and it is used at the core of the open source platform for the Internet of Things, OpenIoT. In this paper we present a Schema Editor that provides an intuitive web interface for defining new types of sensors, and concrete instances of them, using the SSN ontology as the core model. This editor is fully integrated with the OpenIoT platform for generating virtual sensor descriptions and automating their semantic annotation and registration process.

\end{abstract}
\section{Motivation}\label{sec:motivation}

The Internet of Things (IoT) paradigm is expected to dramatically change the way we produce, transmit and process data. IoT makes it possible for devices, objects, people, and \textit{things}, to observe, collect and send all sorts of data in different domains, ranging from environmental sensing to health monitoring or smart cities. As a result, a large number of highly heterogeneous interconnected objects will contribute to the Web of Data, challenging IoT systems to effectively exploit and make use of this data. 
One way to deal with this heterogeneity is through semantic models that provide explicit meaning about the data that is represented. Semantic technologies such as OWL and RDF are  standards for modeling and defining concepts and relationships in arbitrary domains of use, and constitute a promising solution to help coping with this problem.
Based on these well-founded semantic technologies, the OpenIoT open-source platform for IoT (\url{http://openiot.eu/}) provides a flexible cloud-based architecture that helps manage the life cycle of IoT services and applications. The OpenIoT architecture includes, among others, modules that manage the sensor data acquisition, namely X-GSN, the semantic data provision and querying (Linked Sensor Middleware-Light, namely LSM-Light), as well as front-end tools for data discovery and analytics (e.g. Request Definition, and Request Presentation). The integration of all these modules is possible thanks to the use of the OpenIoT ontology, which is based on the SSN ontology~\cite{compton2012}. However, these core ontology models are not specific to any domain, and therefore require to be extended or complemented with other vocabularies in order to be used in practice. 

General purpose ontology editors (e.g. Protégé~\cite{knublauch2004}) are suitable for defining, modifying and customizing ontologies, but they require users to be familiar with ontology modeling and the basics of description logics. Considering that users of IoT platforms are usually not well-versed in ontological engineering, this can represent an overkill for IoT system administrators/users who simply need to add a new sensor or a type of sensor. Moreover, the general purpose editors are not integrated into the workflow of an IoT system (e.g. as OpenIoT) in such a way that sensor descriptions generated are automatically published as Linked Data, and ready to be discovered, queried and re-used. Hence, it is vital to provide simple and intuitive tools that allow IoT users to perform tasks such as add a new sensor or a sensor type intuitively while preserving the ontological foundations of the model.

The Sensor Schema Editor of OpenIoT that we present in this paper aims at providing a solution to this problem. In this first evolution of the editor, we provide the means to: (i) define/modify new sensor types, and (ii) create new sensor instances. A novel feature of the Sensor Schema Editor compared to other UI-based ontology editors \cite{armin2} is that it is a fully functional, implemented prototype completely integrated with the OpenIoT system. The extensions to the ontology generated by the creation of new sensor types are linked dynamically to the OpenIoT ontology using the LSM-Light component. Hence, the extensions to the ontology created are accessible and visible to other system components.

\vspace{-10pt}

\section{Sensor Schema Editor}\label{sec:editor}

\begin{figure}[t!]
	\centering
	\includegraphics[scale=0.8]{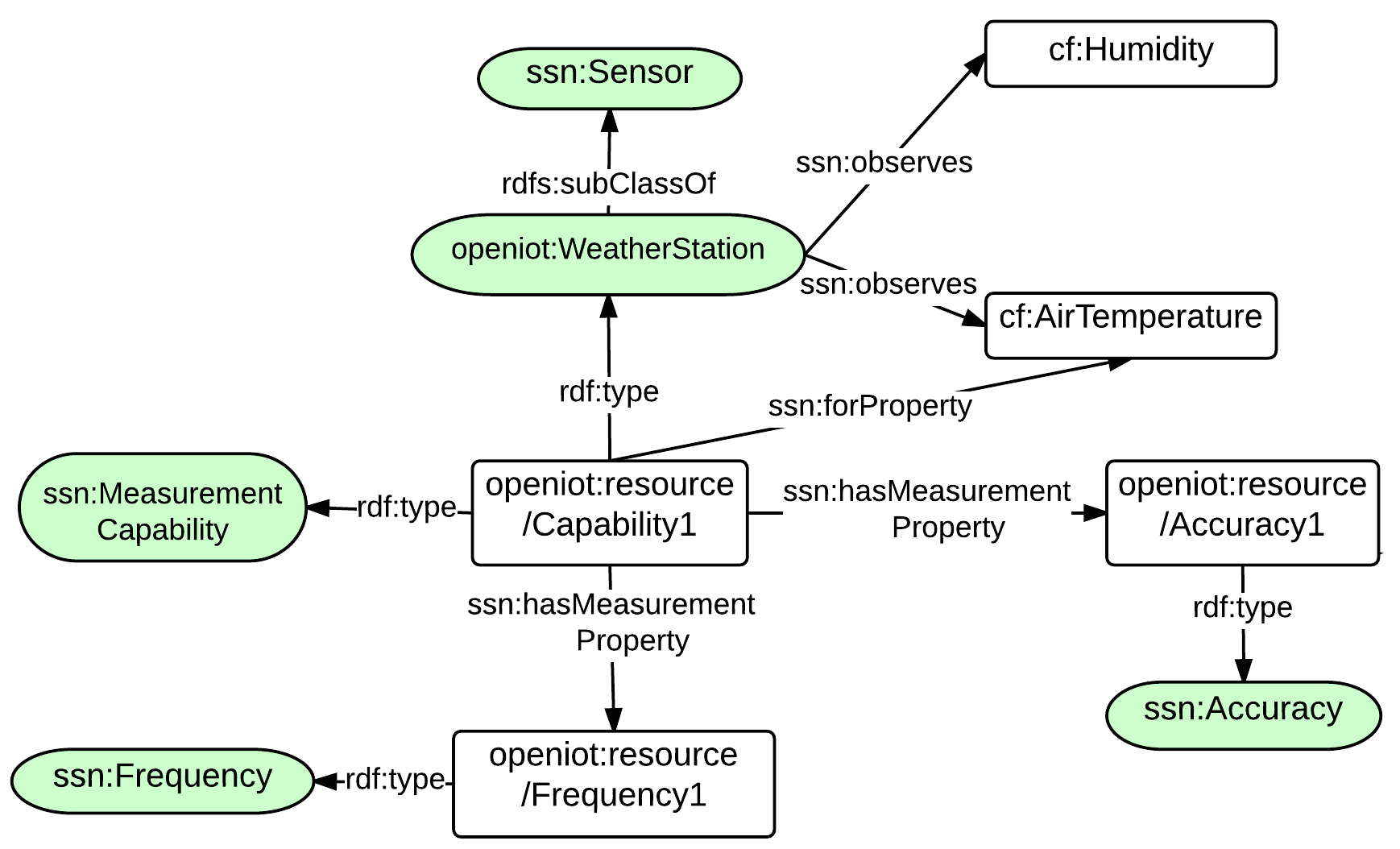}
    \caption{Description of Sensor Types in the OpenIoT Ontology}
    \label{fig:sensortype_definition}
\end{figure}

\begin{figure}[t!]
	\centering
	\includegraphics[scale=0.9]{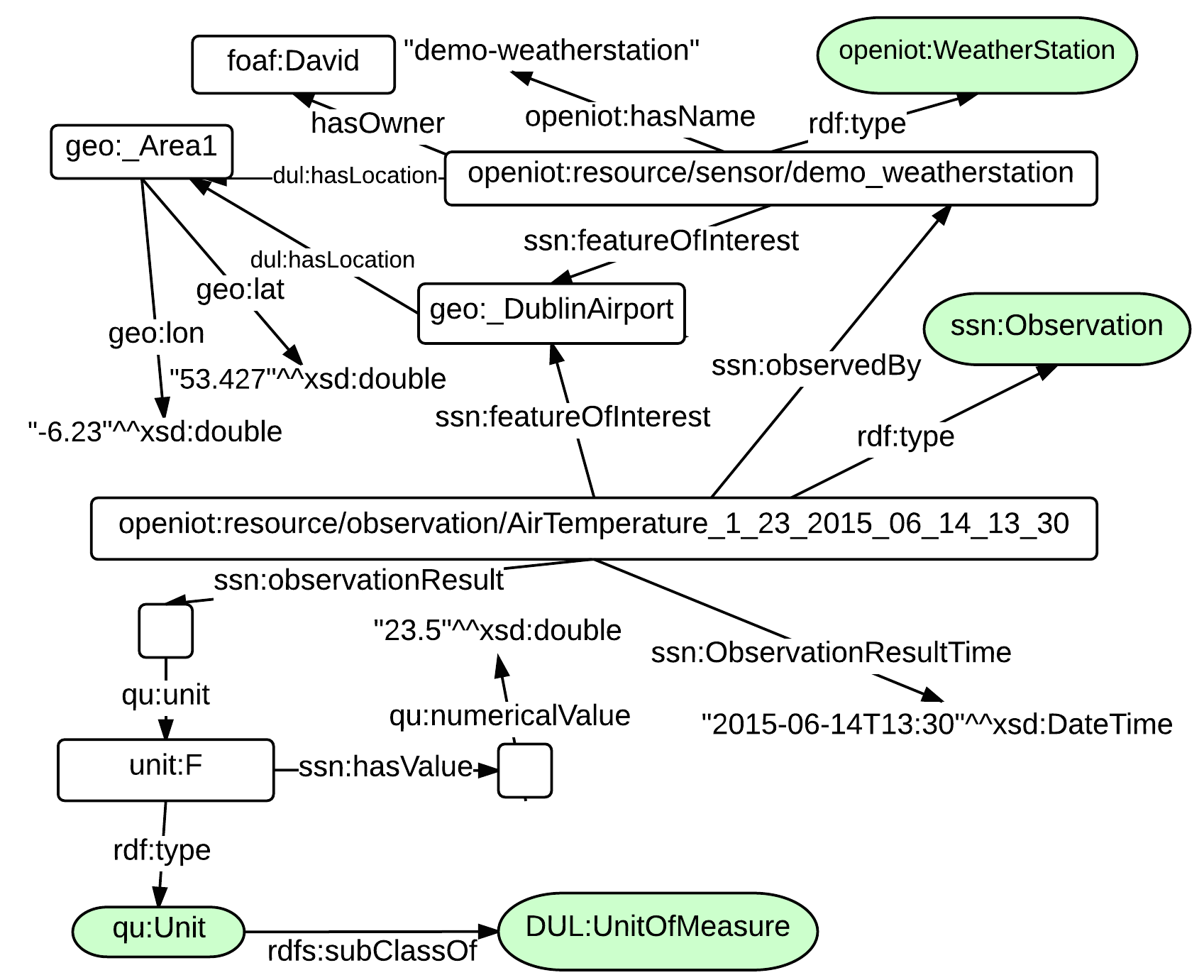}
    \caption{Description of a Sensor Instance in the OpenIoT System}
    \label{fig:sensorinstance}
	\vspace{-10pt}
\end{figure}

The Sensor Schema Editor supports the average user in annotating sensors and sensor-related data using the OpenIoT ontology and Linked Data principles. The interface automates the generation of RDF descriptions for sensor node information submitted by the users. 
As an example, let us consider an IoT deployment where dozens of \texttt{WeatherStation} sensors are deployed in a determined geographical area. In OpenIoT~\cite{soldatos2015}, all sensors and observations are represented in terms of ontological concepts. For example, Figure~\ref{fig:sensortype_definition} depicts a description of a sensor type following the SSN-based OpenIoT ontology. A sensor (e.g. \texttt{WeatherStation}) measures air temperature and humidity, and has some pre-defined \textit{accuracy} and \textit{frequency} parameters, typically defined by the vendor specification or configuration. This sensor type constitutes an extension of the ontology for this particular use case. Based on this new type of sensor, we are able to create instances with user provided descriptions that represent deployed sensors of that type. The LSM-Light component will then semantically annotate and publish the sensor type and instance descriptions as Linked Data, making it searchable and discoverable through SPARQL queries. Figure \ref{fig:sensorinstance} illustrates an overview of how the sensor instance is annotated and published in Linked Data format based on the new sensor type (e.g. \texttt{WeatherStation}) created. The annotation process strictly follows the OpenIoT ontology which is an extension of SSN ontology.


\vspace{-15pt}

\section{Sensor Schema Editor Implementation}\label{sec:editorimplementation}

\vspace{-8pt}

The Sensor Schema Editor\footnote{Available as part of OpenIoT on Github: \url{https://github.com/OpenIotOrg/openiot}} has two components: 1) a web-based interface (Sensor Type and Instance Editors) and 2) a back-end server. The web interface is developed in Java using the JSF framework. The back-end is also developed in Java and employs the Restlet framework (http://restlet.org/). The current implementation of the Sensor Schema Editor is capable of generating new sensor types and instances based on the OpenIoT ontology. 

\vspace{-5pt}

\subsection{Sensor Type Editor}

\begin{figure}[t!]
	\centering
	\includegraphics[scale=0.2]{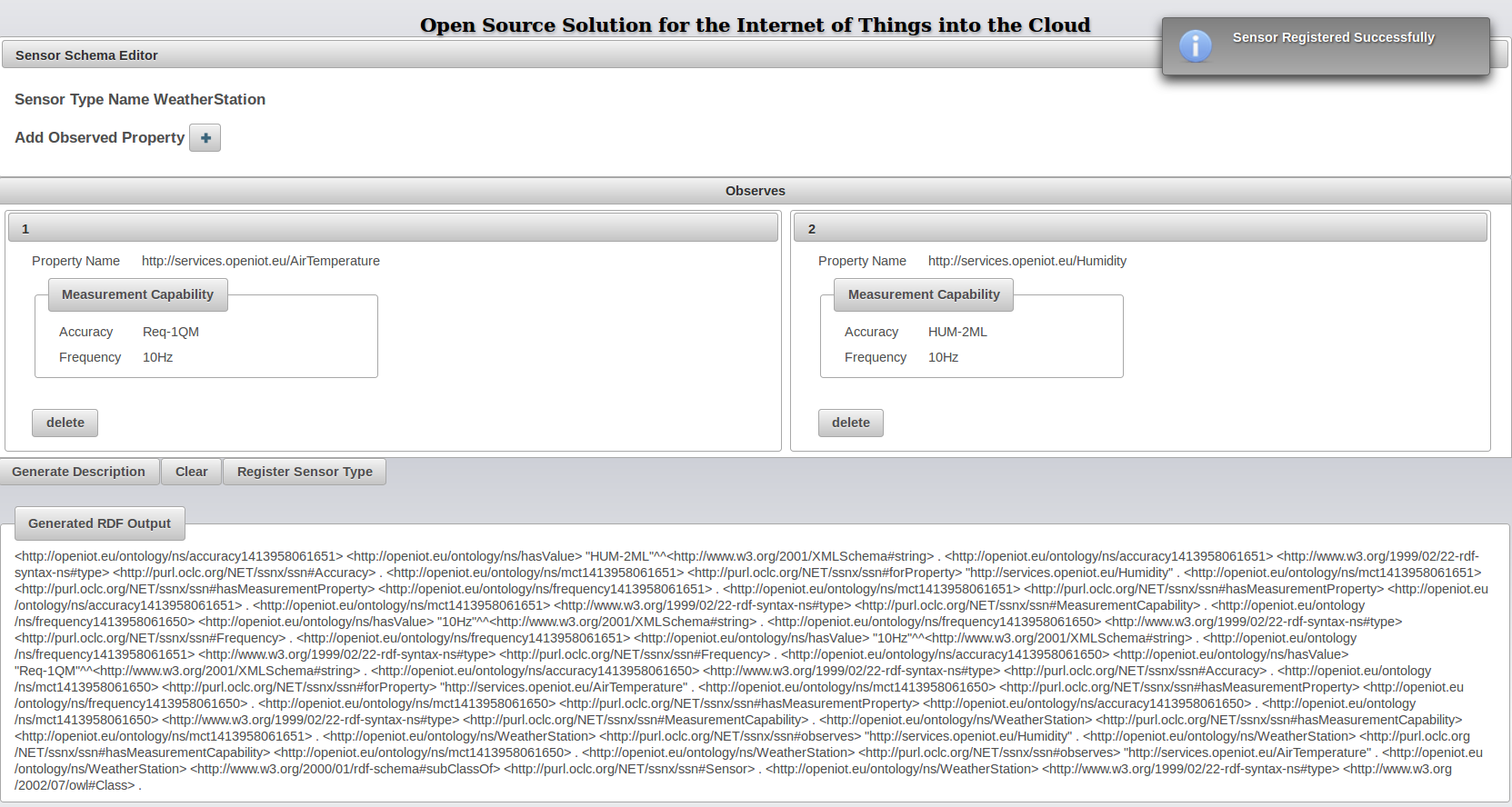}
    \caption{OpenIoT Sensor Schema Editor: Sensor Type Interface}
    \label{fig:editortype}
    \vspace{-15pt}
\end{figure}

Figure \ref{fig:editortype} presents the Sensor Type Editor, an easy to use intuitive interface allowing novice users to define new sensor types. It supports the following concepts to define a new sensor type.

\textit{Sensor Type Name/id}: A human friendly name for the new sensor type.  

\textit{Observed Property:} A property that is observed by the new sensor type. The \textit{observes} relation is used to define the relation between a \textit{sensor} and its \textit{property}. The editor allows a sensor to be associated with multiple observed properties.

\textit{MeasuringCapability:} Collects together measurement properties, in particular the \textit{accuracy} and \textit{frequency}. \textit{Accuracy} is the closeness of agreement between the value of an observation and the true value of the observed quality. \textit{Frequency} is the smallest possible time between one observation and the next.

\textit{Register:} The sensor type editor also provides means to generate the RDF description of the sensor and register it with the OpenIoT LSM-Light service. This allows the sensor type to be discovered, queried and re-used by user communities.

In the example depicted in Figure \ref{fig:editortype}, we define a sensor type \texttt{WeatherStation}. This sensor observes two properties namely \texttt{AirTemperature} and \texttt{Humidity} (URIs). Each of these properties has an associated measurement capability (\textit{accuracy} and \textit{frequency}) that can be defined by the user depending on the datasheet provided by the sensor manufacturer.




\begin{figure}[t!]
    \centering
 \includegraphics[width=0.8\textwidth]{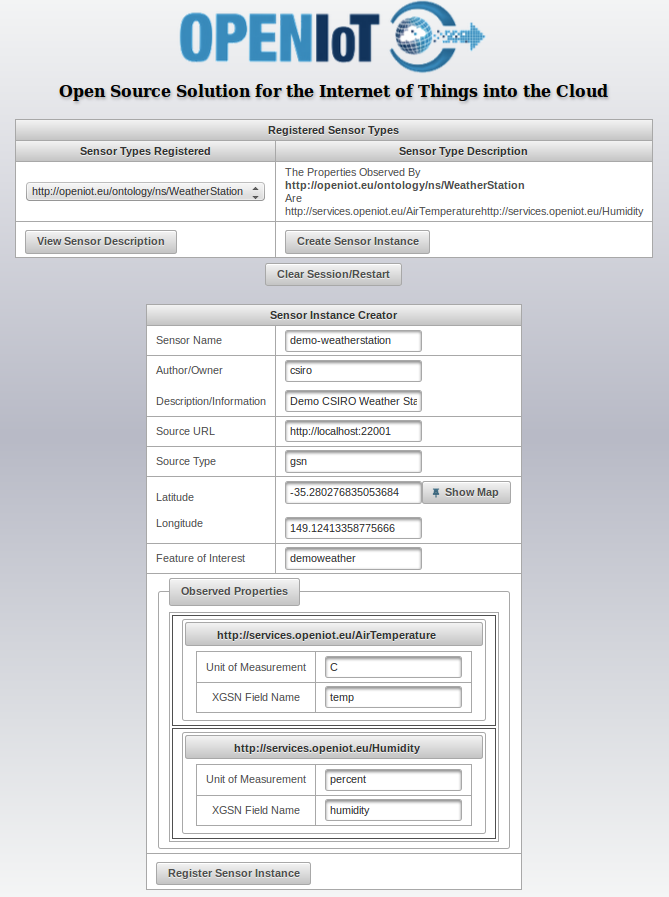}
    \caption{OpenIoT Sensor Schema Editor: Sensor Instance Interface}
    \label{fig:editorinstance}
	\vspace{-10pt}
\end{figure}

\vspace{-5pt}
\subsection{Sensor Instance Editor}

The OpenIoT sensor instance editor uses the sensor type definition created earlier, to generate a concrete (deployed) sensor instance. The instance is a representation of the actual physical/virtual sensor. Figure \ref{fig:editorinstance} provides a screenshot of the sensor instance editor. The instance includes the following information:

\textit{Sensor Name}: the identification of the deployed sensor, e.g. \texttt{demo-weatherstation}

\textit{Owner/Description}: Provides sensor description including who owns it.

\textit{Location:} The physical location of the sensor (based on a Map).

\textit{Feature of Interest:} This is used within the OpenIoT ontology to dynamically link the sensor instance to a domain ontology, e.g. \texttt{demo-weatherstation} points to the observed feature of interest \texttt{crop-growth}. 

\textit{Observed Properties:} These are fetched from the sensor type definition.  The user specifically can define the unit of measurement (e.g. \textit{Kelvin} or \textit{Celsius} for temperature) and the mapping of the ontology observed property field to the X-GSN component (responsible to stream data from sensors). The mapping allows X-GSN to semantically annotate incoming sensor data streams with the sensor instance and type description.

\textit{Generate Metadata:} This function registers the sensor instance with the LSM-Light component and also provides the user with a metadata configuration file required for the functioning of X-GSN.

\vspace{-10pt}

\section{Conclusions}\label{sec:conclusions}
\vspace{-5pt}

In this paper we have presented a web-based Sensor Schema Editor that assists users defining new types of sensors, thus extending the underlying ontology; and creating instances of them in the form of Linked Data, using the SSN ontology as its core model. This editor is part of the OpenIoT open-source platform for IoT development and deployment, and it bridges the gap between the know-how of IoT system administrators, and the SSN-based ontology model that governs the components of OpenIoT. In the future we plan to include customizing other parameters of the sensor description (e.g. custom measurement capabilities) and adding more complex validation mechanisms that alert the user if the produced schemas may produce conflicts in the ontology model. Furthermore, we plan to allow bulk generation of instances for the cases where large numbers of sensor instances need to be created.

\vspace{-10pt}   
\subsubsection*{Acknowledgments}
Supported by the SNSF Nano-Tera OpenSense2 project.

\vspace{-10pt}

\bibliography{rsp}
\bibliographystyle{splncs03}

 \end{document}